\documentclass[twoside,leqno,twocolumn]{article}

\usepackage[letterpaper]{geometry}

\usepackage{ltexpprt}
\usepackage{hyperref}
\usepackage{balance}
\usepackage{graphicx}
\usepackage{amsmath} 
\usepackage{booktabs}
\usepackage{amssymb}
\usepackage{multirow}
\usepackage{xcolor}

\hypersetup{
    colorlinks=true,
    linkcolor=magenta,
    filecolor=magenta,      
    urlcolor=magenta,
    citecolor=cyan,
}
\begin{document}

\title{CoMAL: Collaborative Multi-Agent Large Language Models for Mixed-Autonomy Traffic}

\author{%
    Huaiyuan Yao$^{1}$\thanks{Equal contribution.}, 
    Longchao Da$^{1}$\footnotemark[1], 
    Vishnu Nandam$^1$, 
    Justin Turnau$^1$, \\
    Zhiwei Liu$^2$, 
    Linsey Pang$^2$, 
    Hua Wei$^1$ \\
    $^1$Arizona State University, $^2$Salesforce%
}
\date{}

\maketitle

% Copyright Statement
% When submitting your final paper to a SIAM proceedings, it is requested that you include
% the appropriate copyright in the footer of the paper.  The copyright added should be
% consistent with the copyright selected on the copyright form submitted with the paper.
% Please note that "20XX" should be changed to the year of the meeting.

% Default Copyright Statement
% \fancyhead[L]{THIS IS ON WORK HAS BEEN SUBMITTED TO THE SIAM FOR POSSIBLE PUBLICATION}

% Depending on which copyright you agree to when you sign the copyright form, the copyright
% can be changed to one of the following after commenting out the default copyright statement
% above.

%\fancyfoot[R]{\scriptsize{Copyright \textcopyright\ 20XX\\
%Copyright for this paper is retained by authors}}

%\fancyfoot[R]{\scriptsize{Copyright \textcopyright\ 20XX\\
%Copyright retained by principal author's organization}}

%\pagenumbering{arabic}
%\setcounter{page}{1}%Leave this line commented out.

\begin{abstract} \small\baselineskip=9pt
The integration of autonomous vehicles into urban traffic has great potential to improve efficiency by reducing congestion and optimizing traffic flow systematically. In this paper, we introduce \textbf{CoMAL} (\textbf{Co}llaborative \textbf{M}ulti-\textbf{A}gent \textbf{L}LMs), a framework designed to address the \textit{mixed-autonomy traffic} problem by collaboration among autonomous vehicles to optimize traffic flow. CoMAL is built upon large language models and operates in an interactive traffic simulation environment. Specifically, It utilizes a Perception Module to observe surrounding agents and a Memory Module to store strategies for each agent.
The overall workflow includes a Collaboration Module that encourages autonomous vehicles to discuss the effective strategy and allocate roles, a reasoning engine to determine optimal behaviors based on assigned roles, and an Execution Module that controls vehicle actions using a hybrid approach combining rule-based models. Experimental results demonstrate that CoMAL achieves superior performance on the Flow benchmark. Additionally, we evaluate the impact of different language models and compare our framework with reinforcement learning approaches. It highlights the strong cooperative capability of LLM agents and presents a promising solution to the mixed-autonomy traffic challenge. The code is available at \href{https://github.com/Hyan-Yao/CoMAL}{https://github.com/Hyan-Yao/CoMAL}
\end{abstract}

\section{Introduction}
\begin{figure*}[htb]
  \centering
  \includegraphics[width=\textwidth] {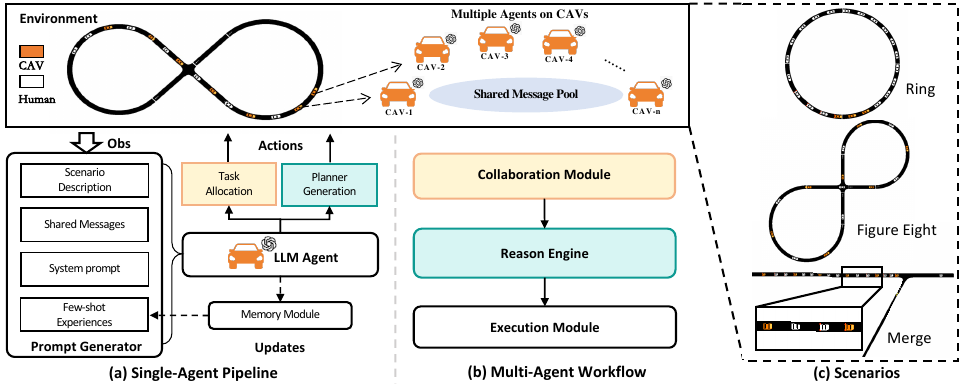}
  \caption{The overall framework of CoMAL. (a) Single-Agent Pipeline: The prompt generator integrates scenario descriptions, few-shot experiences, and shared messages, which are then fed into the LLM. The LLM subsequently allocates tasks and generates planners. (b) Multi-Agent Workflow comprises three modules: the \texttt{Collaboration Module}, the \texttt{Reason Engine}, and the \texttt{Execution Module}. (c) The three Benchmarks Scenarios for CoMAL Ring: The ring road network consists of a closed-loop road where vehicles continuously travel in a circular fashion. {\texttt{Figure Eight (FE)}}: is an extension of the ring road, consisting of two circular loops connected by an intersection. Merge: The merged network simulates how vehicles entering from an on-ramp cause disturbances.}
  \label{fig:pic1}
\end{figure*}

% Background: mixed-autonomy traffic
Recently, there has been significant growth in end-to-end autonomous driving systems~\cite{chib2023recentadvancementsendtoendautonomous}. The integration of large language models (LLMs)~\cite{hallgarten2024vehiclemotionplanninggeneralize, da2024prompt} enhances the ability to generalize to unseen traffic scenarios with embedded common-sense knowledge. These models~\cite{hu2023planningorientedautonomousdriving,dauner2023partingmisconceptionslearningbasedvehicle,mao2023gptdriverlearningdrivegpt} primarily focus on optimizing the performance of individual ego vehicles. However, it also matters to study the problem of \textit{mixed-autonomy traffic} to better deploy autonomous vehicles in society~\cite{pmlr-v87-vinitsky18a}. In mixed-autonomy traffic, connected autonomous vehicles (CAVs) are enabled to collaborate with human-driven vehicles across the traffic network, aiming to optimize overall traffic flow and system-wide efficiency. 

% characteristics & history: complex dynamics -> rule-based, deep RL, shortcomings(generalization, interpretable)
%Traffic systems are extremely complex due to dynamical variance ~\cite{Chang1997AnalysisOC}.
To optimize the overall dynamics in mixed-autonomy traffic, existing researchers propose hand-designed control rules~\cite{7879221, doi:10.1080/00423119508969108,8460567,1246386} to direct a fleet of vehicles to form a desired stable motion pattern. The experiments~\cite{10.1007/978-3-319-10629-8_53, 6588305} suggest that autonomous vehicles can enhance traffic throughput, which highlights the potential of mixed-autonomy systems. To model the complex interactions between autonomous and human-driven vehicles in mixed-autonomy traffic systems, multi-agent reinforcement learning (MARL) methods have proven to be effective in learning cooperative strategies for autonomous vehicles in benchmarking environments like the Flow benchmark~\cite{pmlr-v87-vinitsky18a,wu2021flowmodularlearningframework}.  
%As a deep reinforcement learning (RL) framework, Flow enables the systematic design of RL tasks to improve overall traffic flow, including the control of autonomous vehicles and traffic signals.
While RL-based models demonstrate strong performance in specific controlled mixed-autonomy scenarios~\cite{9770186, liu2024multiagentrolloutapproachhighway}, they showed several weaknesses. Firstly, RL methods struggle to generalize across different scenarios~\cite{DBLP:journals/corr/abs-2112-13112}. This is because in mixed-autonomy traffic, the varied and unpredictable behaviors of human drivers, combining with diverse road networks, makes the training RL rely on large amounts of data. Additionally, it is difficult to understand the decision-making process of RL, which limits its interpretability.

Rethinking human behavioral patterns, adolescents can learn to drive in just 20 hours and handle unfamiliar situations~\cite{LeCun2022APT}, while youngest can spontaneously cooperate to enhance work efficiency~\cite{VELDMAN2020101308}. Human decision-making and cooperation are inherently knowledge-driven, relying on common sense, verbal communication, and reasoning procedures~\cite{wen2024diluknowledgedrivenapproachautonomous}. This is in contrast to data-driven methods that require vast amounts of training data and often struggle with generalization. Recent advancements in LLMs~\cite{Guo2024LargeLM} offer promising knowledge-driven solutions for addressing the mixed-autonomy traffic problem. LLMs can emulate human-like knowledge, including the ability to make decisions, form agreements, analyze road situations, and collaborate in real-time. By leveraging this embodied human knowledge, multiple intelligent LLM agents with distinct roles and attributes can work together to handle complex tasks more efficiently~\cite{talebirad2023multiagentcollaborationharnessingpower}, offering new avenues for improving traffic systems through cooperation and common-sense reasoning.

Building upon these insights and recognizing the limitations of RL in generalizing to new traffic situations, we propose leveraging LLMs for their common-sense reasoning and adaptability. LLMs offer better generalization by drawing on embedded knowledge, allowing them to manage complex and unpredictable traffic dynamics more effectively. We developed an innovative framework named CoMAL specifically designed to address mixed-autonomy traffic challenges, as depicted in Figure 1. CoMAL comprises a simulation environment that enables the interaction of individual LLM agents and allows agents to collect perceptual data from the traffic system. Leveraging the stored experiences within the Memory Module, the agents participate in a brainstorming session in the \texttt{Collaboration Module}, where they allocate tasks and establish their specific roles. Subsequently, each agent develops driving plans according to its designated role through the \texttt{Reason Engine}. Then the Engine generates a rule-based driving planner grounded in the Intelligent Driver Model (IDM), which is then implemented in the \texttt{Execution Module} to compensate for the inherent limitations of LLMs in control performance. Our primary contributions are as follows:

% Our framework: LLM, agents, tool learning

\begin{itemize}
  \item [1)] To the best of our knowledge, this paper is the first to integrate the collaborative capability of multi-agent language models in autonomous driving. The \texttt{Collaboration Module} effectively establishes a session for multiple agents to engage in brainstorming and task allocation.
  \item [2)] We propose CoMAL, a multi-agent framework designed to address mixed-autonomy traffic challenges. CoMAL integrates rule-based planners to enhance control mechanisms while leveraging the cooperative and reasoning ability of LLM.
  % \xlp{I am a little bit confused about Reason Engine and Rule-based planner...Reason Engine is from LLMs reasoning (in the section of 3.2.2), right? then rule-based planner..? I think need to clarify ...?} \hyan{Yes. Reason Engine generates a rule-based IDM planner with various parameters like IDM(v=1.0, a=2.0, s=1.0). IDM is real controller to give acc \xlp{thanks, i see...,now i am clear, maybe add a little bit clear description...}}
  \item [3)] We evaluate CoMAL on \texttt{Flow} benchmark in three classical traffic scenarios as shown in Figure \ref{fig:pic1} (c) and compare its performance against RL methods, with experimental results demonstrating significant performance improvements on average velocity and driving smoothness across various LLMs. 
  \item [4)] We have conducted experiments using multiple LLM models, including GPT-4o-mini and Qwen-72B/32B/7B, showcasing the adaptability of CoMAL across a diverse range of LLMs.
  \end{itemize}
  
%   \begin{figure}[h]
%   \centering
%   \includegraphics[width=0.5\textwidth] {visualization.pdf}
%   \caption{Benchmarks Secnarios for CoMAL: Figure Eight (FE), Ring and Merge. Ring: The ring road network consists of a closed-loop road where vehicles continuously travel in a circular fashion. FE: The figure-eight network is an extension of the ring road, consisting of two circular loops connected by an intersection. Merge: The merge network simulates how vehicles entering from an on-ramp cause disturbances.}
%   \label{fig:scenario}
% \end{figure}

\section{Related Work}

\subsection{Mixed-Autonomy Traffic}
\ \newline
Mixed-autonomy traffic, where connected autonomous vehicles (CAVs) along with human-driven vehicles exist in a system~\cite{pmlr-v87-vinitsky18a}, presents a significant challenge in traffic dynamics modeling and control. A control strategy named the ``slow-in, fast-out" approach~\cite{10.1007/978-3-319-10629-8_53} has demonstrated improvements in traffic throughput with a minimal percentage of autonomous vehicles.

Reinforcement learning (RL) offers a more dynamic and adaptable solution. Benchmarks in RL~\cite{duan2016benchmarkingdeepreinforcementlearning} like Mujoco and the Arcade Learning Environment~\cite{DBLP:journals/corr/abs-1207-4708} provide systematic evaluation and comparison of algorithms. Especially for Mixed-Autonomy, benchmark Flow~\cite{pmlr-v87-vinitsky18a} proposes four traffic scenarios to illustrate distinct RL problems including shockwave minimization, inflow management, efficient merging, and intersection control. It evaluates and compares RL algorithms like Trust Region Policy Optimization (TRPO)~\cite{wu2021flowmodularlearningframework, schulman2017trustregionpolicyoptimization}, Proximal Policy Optimization (PPO)~\cite{schulman2017proximalpolicyoptimizationalgorithms}, Evolutionary Strategies (ES)~\cite{salimans2017evolutionstrategiesscalablealternative}, and Augmented Random Search (ARS)~\cite{mania2018simplerandomsearchprovides} in traffic scenarios.~\cite{mei2023reinforcement} also explores the mixed-autonomy scenario in a multi-agent traffic signal control system~\cite{mei2023libsignal}. However, none of the existing work explores the LLMs' applicability in mixed-autonomy traffic problems.

\subsection{Large Language Model-based Multi-Agents}

Large Language models (LLMs) have become integral to multi-agent systems~\cite{Guo2024LargeLM} due to their capabilities in generalization and common-sense reasoning. LLM-based multi-agent systems leverage these strengths to enhance decision-making and communication among agents~\cite{Wang_2024}. This approach is particularly beneficial in complex scenarios like mixed-autonomy traffic, where it is essential to have effective interaction between human drivers and autonomous systems.

The communication structure of LLM-based multi-agent systems varies across different studies to address specific challenges ~\cite{Guo2024LargeLM, masterman2024landscapeemergingaiagent, talebirad2023multiagentcollaborationharnessingpower}. For example, research~\cite{talebirad2023multiagentcollaborationharnessingpower} has explored both centralized and decentralized communication structures for LLM-based multi-agent systems. In traffic control, LLMs facilitate human-machine interaction and improve decision-making processes. The survey~\cite{Guo2024LargeLM} discusses how LLMs can be integrated into multi-agent systems to enhance communication and coordination among agents in traffic environments. And the study DiLu~\cite{wen2024diluknowledgedrivenapproachautonomous} explores a knowledge-driven approach using LLMs, while LLMLight~\cite{lai2024llmlightlargelanguagemodels} utilizes LLMs as decision-making agents for traffic signal control. A pioneering model 
 Open-TI~\cite{da2023opentiopentrafficintelligence} that integrates LLMs with external traffic analysis tools to perform comprehensive traffic simulations and task-specific operations. Their findings suggest that incorporating domain-specific knowledge into LLMs can significantly enhance the performance and reliability of multi-agent systems in complex scenarios. Even though there exists work about LLM agents for traffic problems, forming better cooperation among agents remains challenging, this paper proposes a novel way to solve the mixed-autonomy task by efficient multi-agent coordinations. 

\begin{figure*}[htb]
  \centering
  \includegraphics[width=\textwidth] {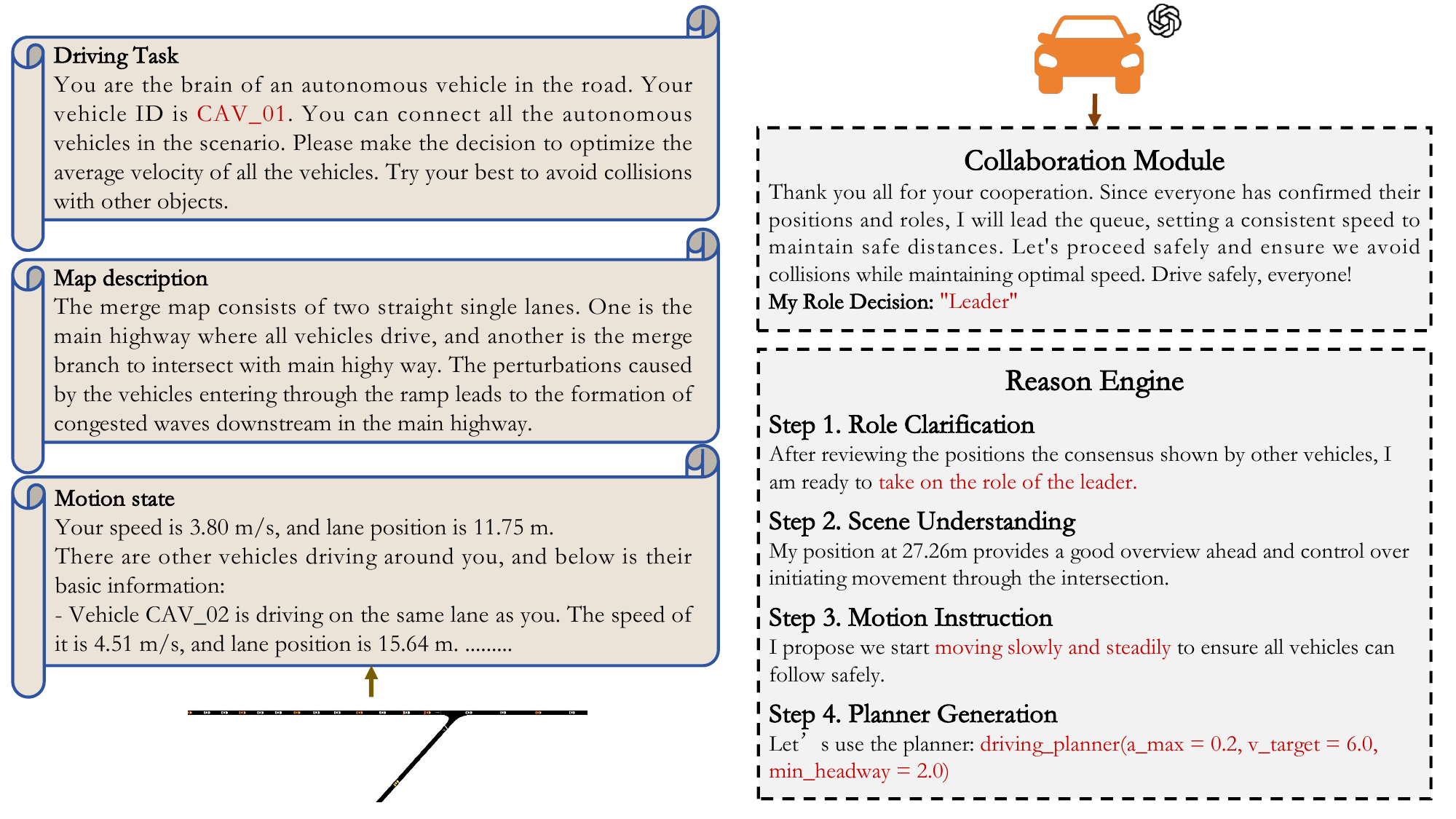}
  \caption{(a) Left: A detailed prompt example for CoMAL, consisting of a system prompt that specifies the driving task, along with map description and motion state provided by the Perception Module. (b) Right: A case of the collaboration and reasoning process. Following task allocation during brainstorming, a hierarchical chain of thought breaks down the driving plan into incremental steps, ensuring consistency in decision-making. This process includes role clarification, scene understanding, motion instruction, and planner generation.}
  \label{fig:prompt}
\end{figure*}

\section{Methodology} 
We introduce CoMAL, a framework designed for LLM agents integrated into connected autonomous vehicles (CAVs) to collaborate and enhance the overall velocity and driving smoothness of traffic flow. As illustrated in Figure~\ref{fig:pic1}, we delineate CoMAL at two distinct levels: the single-agent pipeline and the multi-agent workflow.

At the single-agent level, the LLM-based agents make decisions based on prompts that include few-shot experiences stored in memory, as well as scenario descriptions derived from environmental perception. The agents operate in two modes: task allocation within the \texttt{Collaboration Module} and planner generation within the \texttt{Reason Engine}. The multi-agent workflow consists of three modules: the \texttt{Collaboration Module}, the \texttt{Reason Engine}, and the \texttt{Execution Module}. In the \texttt{Collaboration Module}, CoMAL establishes a shared message pool that facilitates brainstorming and collaborative decision-making among agents. Within this shared space, agents collectively allocate tasks, define their respective roles, and formulate individual driving plans. Each agent then generates a rule-based driving planner in \texttt{Reason Engine}, which is subsequently executed within the \texttt{Execution Module} to ensure coordinated driving behavior and smooth traffic flow.

\subsection{Single-Agent Pipeline}
\ \newline
The quality of prompts significantly influences the output quality of LLM. CoMAL utilizes a prompt generator that integrates all essential information for effective decision-making. The workflow for each individual agent involves several steps: (1) encode the scenario into a textual description within the Perception Module; (2) recall relevant driving experiences from the Memory Module; (3) receive shared messages from other agents through the \texttt{Collaboration Module}; (4) generate prompts and feed it into the LLM; (5) decode the LLM response for task allocation or planning purposes. In this section, we detail the Environment Perception Module and Memory Module.

\subsubsection{Environment Perception Module}
To efficiently extract prompts from complex environmental data and enhance the scene understanding of LLMs, we design an Environment Perception Module. This module extracts key information from the simulation environment and constructs a textual scenario description. The description follows a set of standard rules to generate a thorough representation in natural language. The scene information is divided into two parts: static map and dynamic agents, as shown in Figure~\ref{fig:prompt}.

The static map information represents the scenario type, providing semantic priors for vehicle motion planning. The description of the map helps the LLM intuitively understand the scenario's geometry. The dynamic information describes the motion of the ego vehicle and surrounding agents, which directly influences the planning of vehicles' movement.
\subsubsection{Memory Module}
Similar to human drivers, the agent must make decisions based on reasoning processes that are informed by past driving experiences. To achieve this, we employ a Memory Module that stores experiences from previous driving scenarios and handmade instructions. Initially, the agent is provided with a set of predefined experiences, which the LLM then updates continuously as it engages in reasoning during new situations. This approach allows the agent to refine its decision-making over time, improving its performance in diverse driving contexts. 
% \xlp{Maybe we can add a little bit more technical description of memory module using LLM agent, u pls decide: (1) update frequency: whether the Updates often involve adding new experiences as they occur OR Updates may involve summarizing or abstracting information from multiple experiences OR Frequently updated, often overwriting older information OR Limited capacity, focusing on the most relevant current information. (2)Strategy: Buffer-based updates OR Importance-based updates OR Hierarchical structuring OR incorporate new information using few-shot learning techniques and allows quick adaptation to new situations or domains...
%  }

\begin{figure*}[ht]
  \centering
\includegraphics[width=\textwidth] {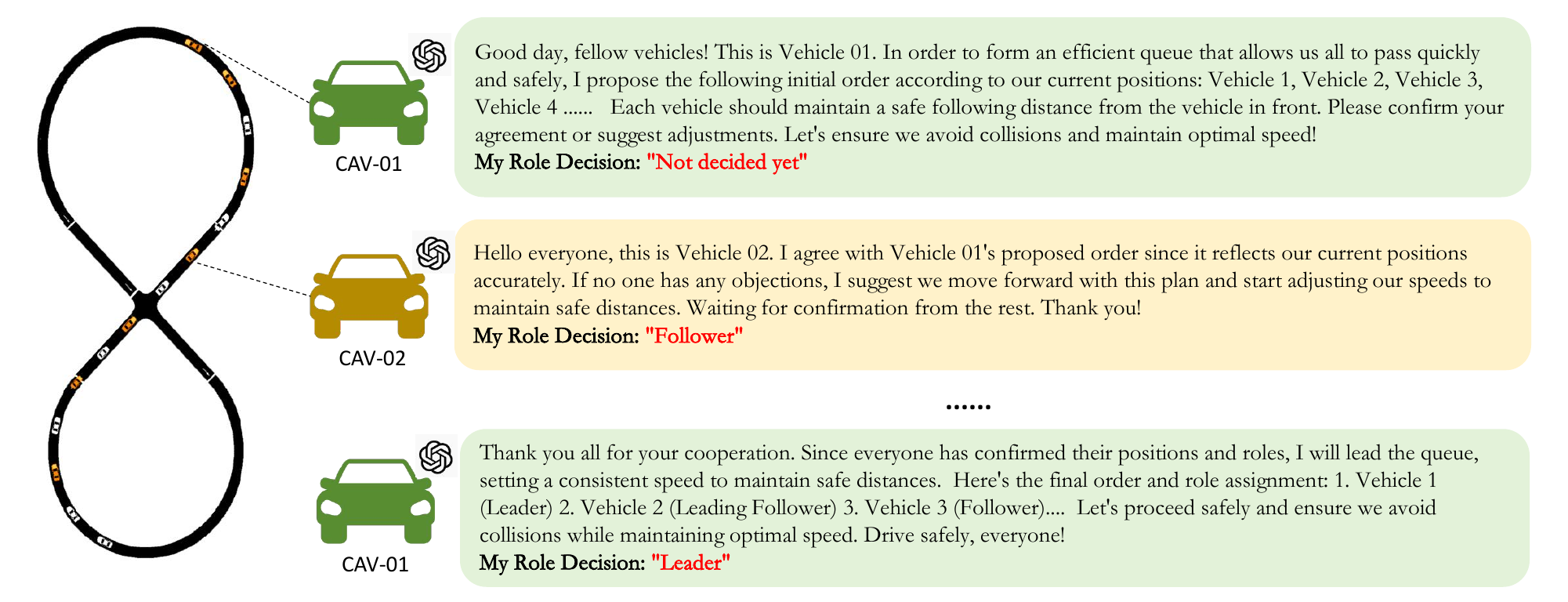}
  \caption{Demonstration of the interaction process of agents in the scenario \texttt{Figure Eight} 1. The agents decide to form a queue and subsequently allocate the roles of leader and follower.}
  \label{fig:collab}
\end{figure*}
\subsection{Multi-Agent Workflow}

In a mixed-autonomy traffic setting, where CAVs operate alongside human drivers, the main objective is to optimize overall traffic flow. To achieve this, we propose a three-stage decision-making workflow. In the \texttt{Collaboration Module}, agents first discuss and allocate tasks within a public message pool. In \texttt{Reason Engine}, each agent then independently determines its driving plan based on its assigned role and generates a driving planner. Finally, the driving planner is executed in the \texttt{Execution Module}.
\subsubsection{Collaboration Module}
Collaborative agents work together towards a shared objective, typically exchanging information to enhance the collective solution. In the \texttt{Collaboration Module}, all CAVs participate interactively by forming a queue for brainstorming and communication. In the brainstorming session, the vehicles take turns subsequently speaking in a public channel to propose strategies and assign tasks among themselves. This collaborative effort ensures that each CAV understands its specific role in the traffic system.

\textbf{Communication Structure} In this section, we introduce a shared message pool to boost communication efficiency, as shown in Figure \ref{fig:collab}. This communication structure maintains a shared message pool where agents can publish messages and subscribe to the latest messages from one another. Agents will take turns to speak one at a time until the strategy is fully developed and each agent's role is clearly defined.

\subsubsection{Reason Engine}
During team brainstorming, each agent determines its role and formulates a strategy to collaborate with other vehicles. Based on these defined roles, the \texttt{Reason Engine} generates an appropriate driving planner to effectively control the vehicle. The \texttt{Reason Engine} takes scenario description and predefined system prompts as inputs. Subsequently, the LLM generates the driving planner based on IDM through reasoning guided by a hierarchical chain-of-thought prompt.

\paragraph{System Prompt} The system prompt defines the planning task and associated driving knowledge. Its primary goal is to standardize the format of both input and output data, as well as clarify the objectives of planner generation. Specifically, it ensures a clear understanding of the physical meaning of each parameter in the IDM planner, such as speed limit ($v_0$), maximum acceleration ($a_m$), and minimum headway ($s_0$). This provides a structured foundation for the decision-making process.

\paragraph{Hierarchical Chain-of-thougts} The hierarchical chain-of-thought process involves four critical components: role clarification, scene understanding, motion instruction, and planner generation. Initially, it is crucial to clarify the role and task of the ego vehicle within a collaborative context. Then the LLM is directed to focus on key information in the scenario, such as headway distance and lead vehicles. Based on the scenario analysis, the LLM is then prompted to provide motion instructions for the ego vehicle. Finally, each agent utilizes scenario analysis and motion instructions to generate a driving planner, parameterized by IDM model~\cite{kesting2010enhanced}.

\subsubsection{Execution Module}
We utilize the rule-based IDM model as a planner to execute driving strategies by adjusting its parameters. IDM is a car-following model to compute longitudinal dynamics. In this model, the acceleration $ a_k $ for vehicle $ k $ is defined by its bumper-to-bumper headway $ s_k $ (distance to preceding vehicle), velocity $v_k$, and relative velocity $\Delta v_k$, via the following equation:

\begin{align}
    a_k = \frac{dv_k}{dt} = a_{\max} [1 - (\frac{v_k}{v_0})^\delta - (\frac{s^*(v_k, \Delta v_k)}{s_k})^2]
\end{align}

where $s^*$ is the desired headway of the vehicle, denoted by:

\begin{align}
    s^*(v_k, \Delta v_k) = s_0 + \max(0, v_k T + \frac{v_k \Delta v_k}{2\sqrt{a_{\max} b}})
\end{align}

where ${s_0, v_0, T, \delta, a_{\max}, b}$ are given parameters. We set the desired time headway $T$, the comfortable braking deceleration $b$, and the acceleration exponent $\delta$ as constants while adjusting the desired velocity $v_0$, the minimum spacing $s_0$, and the maximum acceleration $a_{\max}$ to tailor the driving planners. Thus \texttt{Reason Engine} generates a driving planner by customizing IDM's parameters ${(v_0, a_{\max}, s_0)}$.

%\xlp{might need to have a little more clear description on this section...} \hyan{Updated. May it better!} \xlp{thanks, yes, it is better. then I guess it is also related to Figure 3:Demonstration of prompt. I think these IDM parameters are from the prompt MOtion State value..right? If so, In Figure 3, add some detailed description..}

% combine the table:

\section{Experiments}
In a mixed-autonomy setting, a subset of vehicles are tasked with the objective of improving overall traffic flow and mitigating the formation and propagation of stop-and-go waves. Thus, in our experiments, we aim to address several key questions:

\begin{itemize}
    \item How can CAVs enhance traffic flow and eliminate stop-and-go shockwaves?
    \item How do multiple LLM-based agents collaborate to achieve this goal? 
    \item Do different LLM models influence the results?
\end{itemize}

\subsection{Implementation Details}
The experiments are conducted in
Flow~\cite{wu2021flowmodularlearningframework} with SUMO~\cite{SUMO2018}, a
microscopic simulator for traffic and vehicle dynamics. 
% Flow enables the systematic creation of a variety of traffic-oriented RL tasks for the purpose of generating control strategies for autonomous vehicles, traffic lights, etc. 
For details on the architecture and on training autonomous vehicles to maximize system-level velocity, we refer the readers to~\cite{wu2021flowmodularlearningframework}. The environment offers several driving models to simulate human driver and a realistic interaction between vehicles. We adopt OpenAI GPT-4o-mini, Qwen-72B/32B/7B in this paper.
\subsection{Scenarios}
We evaluate our model on the Figure Eight (FE), Ring, and Merge scenarios from the Flow benchmark. Further details are provided below and illustrated in Figure~\ref{fig:pic1} (c). 
% \paragraph{Ring} The ring road network consists of a circular lane with a specified length.

% \paragraph{Figure Eight (FE)}  The figure-eight network is an extension of the ring road network. It consists of two rings positioned at opposite ends of the network, connected by an intersection. The road segments linking the rings have a length equal to the diameter of the rings. In a mixed-autonomy setting, a portion of vehicles are treated as CAVs with the objective of preventing stop-and-go waves.

% \paragraph{Merge} The merge network highlights the effect of disturbances on vehicles in a highway network. Specifically, perturbations resulting from vehicles arriving from the on-merge lead to the formation of backwards propagating stop-and-go waves, thereby reducing the throughput of vehicles in the network. This phenomenon is known as convective instability. In a mixed-autonomy setting, a percentage of vehicles in the main highway are tasked with the objective of dissipating the formation and propagation of stop-and-go waves from locally observable information. Moreover, given the open nature of the network, the total number of CAVs within the network may vary at any given time.
\paragraph{Ring} The ring road network consists of a circular lane where vehicles continuously travel in a loop. It is commonly used to study traffic dynamics, as disturbances can cause stop-and-go waves. In mixed-autonomy scenarios, CAVs are deployed to reduce these waves and enhance traffic flow stability.

\paragraph{Figure Eight (FE)} The FE network builds on the ring road by connecting two circular loops via an intersection. In mixed-autonomy scenarios, CAVs are introduced to smooth traffic and prevent stop-and-go waves.

\paragraph{Merge} The merged network simulates highway disturbances caused by vehicles entering from an on-ramp, which creates stop-and-go waves. In mixed-autonomy scenarios, CAVs are tasked with mitigating these waves based on local observations and adjusting to fluctuating vehicle numbers in the open network.

We investigate different levels of difficulty for each proposed benchmark by adjusting their scenario-specific meta-parameters. Table~\ref{tab:config} provides detailed descriptions of the selected meta-parameters for each benchmark.

\begin{table}[h!]
\setlength{\tabcolsep}{2pt}
  \caption{Configurations of Benchmarks}
  \label{tab:config}
  \begin{tabular}{lccccl}
    \toprule
    Scenario Name & Time(s) & Vehicles Distribution \\
    \midrule
    FE 0 & 150 & 13 humans, 1 CAV \\
    FE 1 & 150 & 7 humans, 7 CAVs \\
    FE 2 & 150 & 0 humans, 14 CAVs \\
    Ring 0 & 150 & 21 humans, 1 CAV \\
    Ring 1 & 150 & 19 humans, 3 CAVs \\
    Ring 2 & 150 & 11 humans, 11 CAVs \\
    Merge 0 & 75 & 10.0\% CAV penetration rate \\
    Merge 1 & 75 & 25.0\% CAV penetration rate \\
    Merge 2 & 75 & 33.3\% CAV penetration rate \\
    Merge 3 & 75 & 50.0\% CAV penetration rate \\
    Merge 4 & 75 & 90.0\% CAV penetration rate \\
\bottomrule
\end{tabular}
\vspace{-5mm}
\end{table}

\subsection{Metrics} To provide a comprehensive assessment of traffic flow and mitigate the occurrence of shockwaves, we utilize two metrics:
\begin{itemize}
\item Average vehicle speed in the network (m/s). Higher values indicate better overall traffic flow. 
    \item Standard deviation of vehicle speed (m/s). The smaller is more stable. Lower values reflect greater stability and consistency in traffic movement.
\end{itemize}

\subsection{Specification on Communication}
In this section, we focus on the interactive process among agents as they work to solve the mixed-traffic problem. In the FE scenario, the agents recognize the need to form a queue, identify a leader, and designate the remaining agents as followers. The process of task allocation and leader selection is illustrated in Figure \ref{fig:collab}. 
Additionally, in the ring and merge scenarios, agents aim to eliminate shockwaves. Their reasoning is as follows: if there is relative traffic congestion ahead of the ego vehicle, the agent approaches the lead vehicle slowly; otherwise, it accelerates to follow the lead vehicle closely.
\begin{table*}[t!]
\setlength{\tabcolsep}{3pt}
  \caption{Quantitative Evaluation of CoMAL on Flow Benchmarks}
  \label{tab:human}
  \begin{tabular}{cccccccccccccl}
    \toprule
    Metric & Model & FE 0 & FE 1 & FE 2 & Ring 0 & Ring 1 & Ring 2 & Merge 0 & Merge 1 & Merge 2 & Merge 3 & Merge 4 \\
    \midrule
    \multirow{2}{*}{Avg} & Human Driver & 5.61 & 5.61 & 5.61 & \textbf{2.88} & \textbf{2.88} & \textbf{2.88} & 6.40 & 6.40 & 6.40 & 6.40 & 6.40  \\
     & \textbf{CoMAL} & \textbf{6.40} & \textbf{6.47} & 
\textbf{6.29} & 2.86 & 2.85 & 2.87 & \textbf{6.59} & \textbf{7.40} & \textbf{7.42} & \textbf{7.86} & \textbf{8.83} \\
    \midrule
    \multirow{2}{*}{Std} & Human Driver & 4.55 & 4.55 & 4.55 & 0.79 & 0.79 & 0.79 & 3.12 & 3.12 & 3.12 & 3.12 & 3.12  \\
     & \textbf{CoMAL} & \textbf{1.74} & \textbf{1.77} & \textbf{2.24} & \textbf{0.29} & \textbf{0.26} & \textbf{0.31} & \textbf{2.88} & \textbf{2.91} & \textbf{2.61} & \textbf{2.47} & \textbf{2.70} \\
  \bottomrule
\end{tabular}
\end{table*}

\begin{figure*}[t!]
  \centering
  \includegraphics[width=\textwidth] {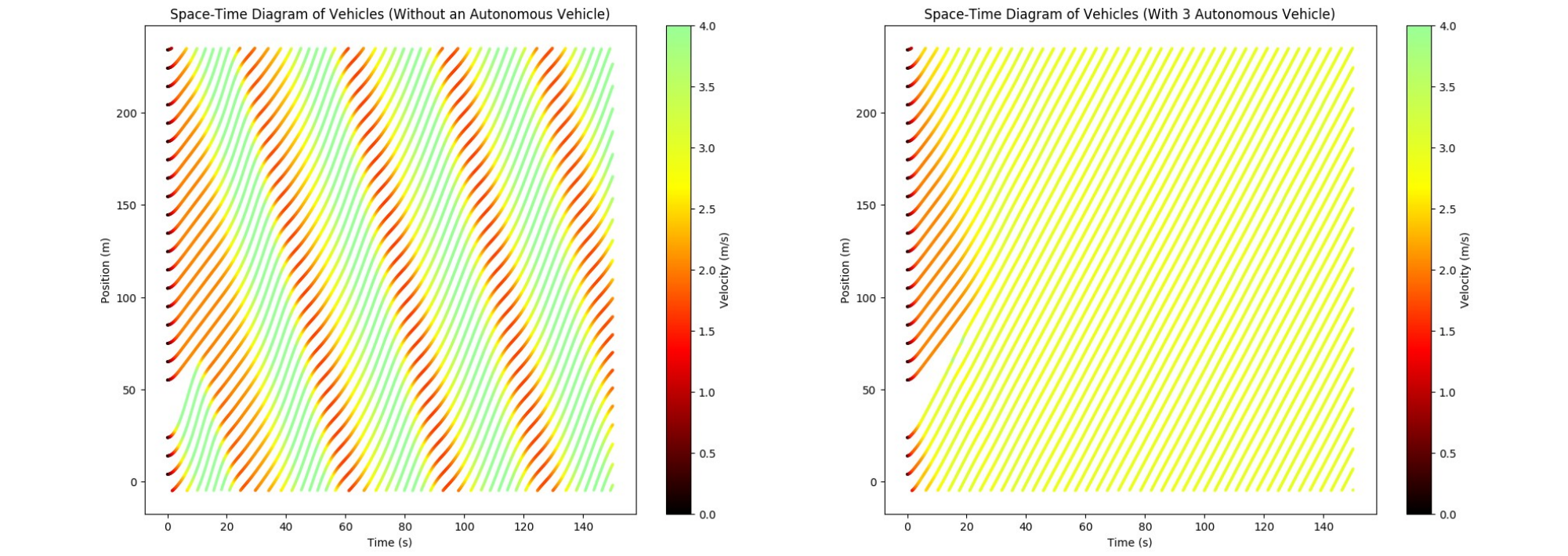}
  \caption{Visualization of vehicle trajectories in Ring 0 setting. The ring road has a total length of 230 meters and contains 22 vehicles. Each line in the space-time diagrams shows the position of a specific vehicle over time, whose speed is indicated with different colors. When a vehicle completes a full lap of the ring, its position resets to zero. \textbf{Left:} In the absence of automated vehicles, human-driven vehicles exhibit stop-and-go shockwaves due to inherent instability. \textbf{Right:} With three connected autonomous vehicles using the CoMAL framework, the unstable vehicles are stabilized.}
  \label{fig:pic2}
  \vspace{-3mm}
\end{figure*}

\subsection{Quantitative Results}
We evaluated our model on the aforementioned benchmarks, varying the number or percentage of CAVs across different settings.

As shown in Table~\ref{tab:human}, we compared the performance of CoMAL with that of human drivers. The results indicate that as the number of CAVs increases, CoMAL's performance generally improves and surpasses that of human drivers. This highlights the strong capability of LLM agents in achieving effective cooperation. A visualization of the vehicle trajectories in Ring 0 setting is shown in Figure~\ref{fig:pic2}. We can see that the proposed CoMAL framework can stabilize the unstable vehicle flow. 

\begin{table}[t!]
\setlength{\tabcolsep}{2pt}
  \caption{Ablation Study}
  \begin{tabular}{cccccl}
    \toprule
    No. & Perception & Memory & Collaboration & FE 1 & Merge 1 \\
    \midrule
    1 & $\times$ & $\times$ & $\times$ & 5.61 & 6.40 \\
    2 & $\times$ & $\checkmark$ & $\checkmark$ & 5.81 & 6.51 \\
    3 & $\checkmark$ & $\times$ & $\checkmark$ & 5.17 &  6.72 \\
    4 & $\checkmark$ & $\checkmark$ & $\times$ & 5.18 & 6.88 \\
    5 & $\checkmark$ & $\checkmark$ & $\checkmark$ & \textbf{6.47} & \textbf{7.40} \\
  \bottomrule
\end{tabular}
\label{tab:ablation}
\end{table}

\begin{table*}[h!]
\setlength{\tabcolsep}{4.5pt}
  \caption{Quantitative Experiment of Different LLMs: Average Velocity and Standard Deviation Analysis} 
  \label{tab:combined}
  \begin{tabular}{lcccccccccccl}
    \toprule
    \multirow{2}{*}{Model} & \multicolumn{10}{c}{Average Velocity (m/s)} \\
    \cmidrule{2-12}
    & FE 0 & FE 1 & FE 2 & Ring 0 & Ring 1 & Ring 2 & Merge 0 & Merge 1 & Merge 2 & Merge 3 & Merge 4 \\
    \midrule
    Human Driver & 5.61 & 5.61 & 5.61 & 2.88 & 2.88 & 2.88 & 6.40 & 6.40 & 6.40 & 6.40 & 6.40  \\
    GPT-4o-mini & 
    \textbf{6.40} & \textbf{6.47} & 
    \textbf{6.29} & 2.86 & 2.85 & 2.87 & 6.59 & \textbf{7.40} & 7.42 & 7.86 & \textbf{8.83} \\
    Qwen-72B & 6.37 & 5.96 & 6.07 & 2.86 & 2.83 & 2.85 & \textbf{6.66} & 7.38 & \textbf{7.46} & \textbf{8.07} & 8.75 \\
    Qwen-32B & 6.39 & 5.73 & 6.11 & 2.84 & 2.86 & 2.81 & 6.58 & 7.12 & 7.39 & 7.73 & 8.54 \\
    Qwen-7B & 5.60 & 5.17 & 4.80 & 2.82 & 2.84 & 2.83 & 6.55 & 6.87 & 7.05 & 7.47 & 8.53 \\
    \midrule
    \midrule
    \multirow{2}{*}{Model} & \multicolumn{10}{c}{Standard Deviation (m/s)} \\ \cmidrule{2-12}
    & FE 0 & FE 1 & FE 2 & Ring 0 & Ring 1 & Ring 2 & Merge 0 & Merge 1 & Merge 2 & Merge 3 & Merge 4 \\
    \midrule
    Human Driver & 4.55 & 4.55 & 4.55 & 0.79 & 0.79 & 0.79 & 3.12 & 3.12 & 3.12 & 3.12 & 3.12  \\
    GPT-4o-mini & 1.74 & \textbf{1.77} & 2.24 & \textbf{0.29} & \textbf{0.26} & 0.31 & 2.88 & 2.91 & 2.61 & \textbf{2.47} & 2.70 \\
    Qwen-72B & \textbf{1.73} & 2.76 & 1.92 & 0.32 & 0.32 & \textbf{0.30} & 2.82 & \textbf{2.71} & \textbf{2.54} & 2.61 & 2.78 \\
    Qwen-32B & 1.74 & 2.37 & 2.64 & 0.33 & 0.37 & 0.33 & \textbf{2.81} & 2.99 & 2.62 & 2.57 & \textbf{2.50} \\
    Qwen-7B & 4.51 & 1.89 & \textbf{1.76} & 0.49 & 0.61 & 0.48 & 3.02 & 2.88 & 2.74 & 2.71 & 2.64 \\
    \bottomrule
  \end{tabular}
\end{table*}

\subsection{Ablation Studies}
We further conducted the detailed ablation analysis of the effectiveness of each component of CoMAL in the Figure Eight (FE) 1 and Merge 1 scenarios, the results are presented in Table \ref{tab:ablation}.

\paragraph{Ablation on Perception}
The comparisons in the second and fifth rows of Table~\ref{tab:ablation} demonstrate the effectiveness of incorporating textual descriptions of the map and agents' motion states in the Perception Module. Once perception information is lost, agents are no longer able to comprehend the spatial relationships between the ego vehicle and surrounding agents. As a result, their capacity for effective collaboration and reasoning is significantly impaired.

\paragraph{Ablation on Memory}
The comparisons in the third and fifth rows of Table~\ref{tab:ablation} illustrate the impact of the \texttt{Memory Module}, in which specific experiences are allocated for each scenario. In the absence of high-quality experiences, agents are more susceptible to errors in both discussion and reasoning thus causing a performance decrease in the overall optimization.

\paragraph{Ablation on Collaboration}
The comparison presented in the fourth and fifth rows of Table~\ref{tab:ablation}, as well as in the first row, highlights the effectiveness of the \texttt{Collaboration Module}. The absence of this module causes all agents to adopt nearly identical strategies in our experiment, which in turn leads to conflicts and duplicated efforts in execution. The absence of such collaboration results in performance that can be even worse than that of the simple rule-based model shown in the first row.

\subsection{Discussion}

\paragraph{Comparison with RL methods:}

We conducted experiments on FE and the Merge scenarios and compared with RL methods developed in~\cite{pmlr-v87-vinitsky18a}. The results are shown in Table~\ref{tab:rl}. The Ring is not included in this experiment because the adopted RL benchmark doesn't have the Ring scenario embedded. In the FE scenario, CoMAL demonstrates robust global collaboration, whereas multi-agent RL models struggle to differentiate roles, hindering effective cooperation. Consequently, CoMAL outperforms RL-based approaches. However, in the Merge scenario, CoMAL performs less effectively than RL, indicating that the collaboration is not global. This finding highlights the critical importance of cooperation in enhancing performance.

\begin{table}[t!]
\setlength{\tabcolsep}{3pt}
  \caption{Comparison to RL Benchmark}
  \label{tab:freq}
  \begin{tabular}{lccccccl}
    \toprule
    Model & FE 0 & FE 1 & FE 2 & Merge 0 & Merge 1 & Merge 2  \\
    \midrule
    IDM & 5.61 & 5.61 & 5.61 & 6.40 & 6.40 & 6.40  \\
    ARS & 7.31 & 6.43 & 5.70 & 11.30 & 11.06 & 11.50 \\
    ES & \textbf{6.87} & - & 5.96 & 13.31 & \textbf{17.29} & \textbf{17.36} \\
    TRPO & 8.26 & 5.61 & 5.03 & \textbf{14.95} & 13.74 & 14.14  \\
    PPO & 8.26 & 5.61 & 5.03 & 13.66 & 14.61 & 14.54 \\
    CoMAL & 6.40 & \textbf{6.47} & \textbf{6.29} & 6.59 & 7.40 & 7.42 \\
  \bottomrule
\end{tabular}
\label{tab:rl}
\end{table}

\paragraph{Comparison of various LLM models:}
We evaluate performance across LLM models of varying sizes (see Table~\ref{tab:combined}). The GPT-4o-mini achieves the highest performance among these. Among open-source models, the Qwen 72B has a similar level with the GPT-4o-mini, while the Qwen 32B shows slightly lower performance, and the Qwen 7B performs significantly worse. Notably, we observe that in scenarios requiring extensive collaboration, the performance of smaller models deteriorates more rapidly. This finding suggests that collaboration is a more challenging task than reasoning within the CoMAL framework.

\section{Conclusion}
In this paper, we present the CoMAL, an effective LLM-based multi-agent framework to address mixed-autonomy traffic challenges. By prompt-tuning LLMs with a hierarchical LLM-based planner, CoMAL is able to handle complex vehicle driving tasks towards a collaborative goal under mixed-autonomy traffic. The LLM agent does so by serving primarily as a high-level commander, coordinating with lower-level controllers to execute detailed operations. We hope this work can demonstrate the significant potential of multi-agent systems driven by LLMs to make informed decisions and collaborate effectively in mixed-autonomy scenarios. 

Besides the great potential, we also acknowledge the limitations of our current work and would like to point out several important future directions. First, a possible extension would be increasing the experimental scale to more agents, and improving their collaboration to observe if emerging behaviors would be formed, just like in RL methods. Second, our paper addresses such mix-autonomy problem by the simplified use of LLMs alone, whereas the combination of RL with LLM might be helpful to take advantage of both: exploration and contextual information, and thus improve the performance on this task. In the future, more sophisticated scenarios will be explored to test the possibility of LLM's behavior in complex tasks. 

\section*{Acknowledgements}
The work was partially supported by NSF awards \#2421839, NAIRR \#240120. This work used AWS through the CloudBank project, which is supported by National Science Foundation grant \#1925001. The views and conclusions contained in this paper are those of the authors and should not be interpreted as representing any funding agencies. We thank OpenAI for providing us with API credits under the Researcher Access program and Amazon Research Awards.

% \printbibliography
% \bibliography{main}

\end{document}